\documentclass{article}

\usepackage{arxiv}
\usepackage[utf8]{inputenc}
\usepackage[T1]{fontenc}
\usepackage[ngerman]{babel}
\usepackage{graphicx}
\usepackage{url}
\usepackage{hyperref}
\usepackage{booktabs}
\usepackage{amssymb,amsmath}
\usepackage{tabularx}
\usepackage{orcidlink}
\usepackage{csquotes}

\title{Benchmarking von ASR-Modellen im deutschen medizinischen Kontext: Eine Leistungsanalyse anhand von Anamnesegesprächen}
\renewcommand{\shorttitle}{Benchmarking von ASR-Modellen für medizinische Kontexte}

\author{Thomas Schuster \orcidlink{0000-0002-9539-1627} \\
    Hochschule Pforzheim \\
    \texttt{thomas.schuster@hs-pforzheim.de}
    \And
    Julius Trögele \orcidlink{0009-0007-5959-4287} \\
    XPACE GmbH \\
    \texttt{julius.troegele@xpace.de}
    \And
    Nico Döring \orcidlink{0009-0004-2018-5417} \\
    XPACE GmbH \\
    \texttt{nico.doering@xpace.de}
    \And 
    Robin Krüger \orcidlink{0009-0004-9937-7437} \\
    XPACE GmbH \\
    \texttt{robin.krueger@xpace.de}
    \And 
    Matthieu Hoffmann \orcidlink{0009-0007-9320-0069} \\
    XPACE GmbH \\
    \texttt{mathieu.hoffmann@xpace.de}
    \And 
    Holger Friedrich \\
    XPACE GmbH \\
    \texttt{holger.friedrich@xpace.de}
}

\date{January 2026}

\begin{document}
\maketitle

\begin{abstract}
Die automatisierte Spracherkennung (Automatic Speech Recognition, ASR) bietet großes Potenzial zur Entlastung von medizinischem Personal, beispielsweise durch die Automatisierung von Dokumentationsaufgaben. Während für die englische Sprache zahlreiche Benchmarks existieren, fehlt es für den deutschsprachigen, medizinischen Kontext noch immer an spezifischen Evaluationen, insbesondere unter Berücksichtigung von Dialekten und Akzenten. In diesem Artikel präsentieren wir einen kuratierten Datensatz von simulierten Arzt-Patienten-Gesprächen und evaluieren insgesamt 29 verschiedene ASR-Modelle. Das Testfeld umfasst sowohl Open-Weights-Modelle der Whisper-, Voxtral und Wav2Vec2-Familien als auch kommerzielle State-of-the-Art APIs (AssemblyAI, Deepgram). Zur Evaluation nutzen wir drei verschiedene Metriken (WER, CER, BLEU) und geben einen Ausblick auf die qualitative semantische Analyse. Die Ergebnisse zeigen signifikante Leistungsunterschiede zwischen den Modellen: Während die besten Systeme bereits sehr gute Word Error Rates (WER) von teilweise unter 3 \% erreichen, liegen die Fehlerraten anderer Modelle, insbesondere bei medizinischer Terminologie oder dialektgeprägten Varianten, deutlich höher.
\end{abstract}



\section{Einleitung}
Formelle und organisatorische Aufgaben, wie die klinische Dokumentation, binden einen erheblichen Teil der Arbeitszeit von Pflegepersonal, Ärztinnen und Ärzten. Sprachtechnologien versprechen hier Abhilfe, indem sie Arzt-Patienten-Gespräche (beispielsweise zur Anamnese) automatisch transkribieren und strukturieren. Die Anforderungen an solche Systeme sind im deutschsprachigen Raum besonders hoch und untergliedern sich unter anderem in diese Bereiche:
\begin{itemize}
    \item \textbf{Medizinische Terminologie:} Ein Mix aus lateinischen, griechischen und deutschen Fachbegriffen.
    \item \textbf{Sprecherzuordnung (Diariesierung):} Die korrekte Zuordnung von Aussagen zu Arzt oder Patient.
    \item \textbf{Akzente und Dialekte:} In der Praxis sprechen Patienten selten Hochdeutsch, und auch das medizinische Personal weist eine hohe Diversität an Sprachhintergründen auf.
\end{itemize}
Ziel dieser Arbeit ist es, die Leistungsfähigkeit moderner ASR-Systeme quantitativ zu bewerten und ihre Eignung für den klinischen Alltag zu prüfen.

\section{Datensatz}
Um die Untersuchung durchzuführen, haben wir zunächst einen neuen Datensatz (\enquote{Med-De-Anamnese}) erstellt, der vier repräsentative Szenarien einer medizinischen Aufnahmeuntersuchung abbildet. Die Daten wurden aus öffentlich zugänglichen Gesprächsdialogen von Youtube bezogen und anschließend manuell transkribiert. Diese manuelle Transkription (Ground Truth) dient uns anschließend zur Evaluation der automatischen Transkription, die durch die Sprachmodelle erstellt werden.

\subsection{Szenarien}
Unser Datensatz umfasst folgende vier Fälle, die verschiedene medizinische Fachbereiche und akustische Herausforderungen abdecken:
\begin{enumerate}
    \item \textbf{Rückenschmerzen (Standard):} Ein klassisches Anamnesegespräch (Verdacht auf Prolaps) mit klarer Artikulation. Dieses Szenario dient als Baseline für die allgemeine Erkennungsleistung.\footnote{Quelle: \url{https://www.youtube.com/watch?v=Vv_eHFeKkBw}}
    \item \textbf{Unterbauchschmerzen (Divertikulitis):} Ein Fall mit spezifischerer internistischer Fachterminologie, der den Wortschatz der Modelle prüft.\footnote{Quelle: \url{https://www.youtube.com/watch?v=9mGj_ASu328}}
    \item \textbf{Ausländische Ärztin:} Ein Szenario (Verdacht auf Beinvenenthrombose), in dem die behandelnde Ärztin einen nicht-muttersprachlichen Akzent aufweist. Dies testet die Robustheit der Modelle gegenüber der im medizinischen Personal häufigen Diversität.\footnote{Quelle: \url{https://www.youtube.com/watch?v=u3vLSuzMDvg}}
    \item \textbf{Starker Akzent (Patient):} Ein komplexes Szenario (Morbus Fabry), bei dem der Patient einen starken regionalen Dialekt und eine undeutliche Aussprache aufweist. Dies stellt die höchsten Anforderungen an die akustische Modellierung.\footnote{Quelle: \url{https://www.youtube.com/watch?v=6NwEyPAvGIA}}
\end{enumerate}

\subsection{Ground Truth Erstellung}
Für jedes Szenario wurden die Texte transkribiert. Dabei wurden zwei Arten von Transkript als Ground-Truth-Daten erstellt:
\begin{itemize}
    \item \textbf{Volltext:} Der reine gesprochene Text, normalisiert (Zahlen ausgeschrieben, Interpunktion geglättet).
    \item \textbf{Sprechererkennung:} Eine als attributiert, segmentierte JSON-Datei, die jeden Sprecherwechsel (Sprecher A vs. Sprecher B) mit Textinhalt festhält, um die Unterscheidung der Sprecher durch die Modelle (Diarization-Leistung) zu messen.
\end{itemize}

\section{Methodik und Evaluationssetup}
In diesem Kapitel beschreiben wir den systematischen Ansatz zur Bewertung der ASR-Systeme. Wir erläutern zunächst die Kriterien für die Auswahl der Modelle, bevor wir die technische Pipeline für die Inferenz und die verwendeten Metriken zur Qualitätsmessung im Detail vorstellen.

\subsection{Modellauswahl und Kategorisierung}
Die Auswahl der evaluierten ASR-Systeme (Automatic Speech Recognition) orientiert sich an aktuellen State-of-the-Art-Architekturen sowie deren Relevanz für den praktischen Einsatz im deutschsprachigen Raum. Um eine umfassende Leistungsmatrix zu erstellen, wurden die Modelle in zwei Primärkategorien unterteilt (siehe Tabelle \ref{tab:models}):

\begin{enumerate}
    \item \textbf{Multilinguale und domänenspezifische Modelle:} Diese Gruppe umfasst Architekturen, die explizit auf deutschen Sprachdaten trainiert oder feinjustiert wurden. Hierunter fallen neben den populären \textit{Whisper}-Varianten \cite{radford2022robust} auch spezialisierte Modelle wie \textit{Wav2Vec2 XLS-R} \cite{XLSRSelfsupervisedSpeech, babu22_interspeech} oder \textit{WhisperX} \cite{li2023whisperx}.
    \item \textbf{Monolinguale englischsprachige Referenzmodelle:} Bewusst wurden auch rein englischsprachige Checkpoints in die Evaluation einbezogen. Erwartungsgemäß führen diese bei deutscher Spracheingabe zu einer signifikanten Degradierung der Erkennungsrate. Ihre Einbeziehung dient jedoch der Bestimmung einer \textit{Performance-Baseline} sowie der Analyse des Zero-Shot-Verhaltens generischer Spracharchitekturen bei phonetisch ähnlichen, aber lexikalisch fremden Eingaben \cite{baevski2020wav2vec, conneau21_interspeech}.
\end{enumerate}

\subsection{Evaluationspipeline}
Um eine objektive Vergleichbarkeit zu gewährleisten, wurde eine generische Inferenz-Pipeline implementiert. Diese stellt sicher, dass alle Modelle unter identischen Bedingungen evaluiert werden:
\begin{itemize}
    \item \textbf{Preprocessing:} Resampling der Audio-Signale auf 16\,kHz sowie eine Lautstärkenormalisierung (Peak-Normalization), um eine konsistente Eingabevarianz über alle Test-Samples zu gewährleisten.
    \item \textbf{Inferenz:} Verwendung einer standardisierten Dekodierungs-Logik auf Basis von \textit{Greedy Decoding}. Dies isoliert die akustische Leistungsfähigkeit des Modells von algorithmischen Optimierungen durch Suchverfahren wie Beam Search.
    \item \textbf{Postprocessing:} Text-Normalisierung durch \textit{Lowercasing} und Stripping von Interpunktion. Dies stellt sicher, dass die Berechnung der \textit{Word Error Rate} (WER, Wortfehlerrate) und \textit{Character Error Rate} (CER, Zeichenfehlerrate) rein die lexikalische Erkennungsleistung widerspiegelt.
\end{itemize}

\begin{table}[ht]
\centering
\caption{Übersicht der evaluierten ASR-Modelle}
\label{tab:models}
\resizebox{\textwidth}{!}{%
\begin{tabular}{@{}lllll@{}}
\toprule
\textbf{Modell} & \textbf{Checkpoint / Version} & \textbf{Typ} & \textbf{Implementierung} & \textbf{Lizenz} \\ \midrule
\multicolumn{5}{l}{\textit{Multilinguale \& Deutsche Modelle}} \\
\midrule
Whisper (tiny) & \href{https://huggingface.co/openai/whisper-tiny}{openai/whisper-tiny} & Lokal & Hugging Face & MIT \\
Whisper (base) & \href{https://huggingface.co/openai/whisper-base}{openai/whisper-base} & Lokal & Hugging Face & MIT \\
Whisper (small) & \href{https://huggingface.co/openai/whisper-small}{openai/whisper-small} & Lokal & Hugging Face & MIT \\
Whisper (medium) & \href{https://huggingface.co/openai/whisper-medium}{openai/whisper-medium} & Lokal & Hugging Face & MIT \\
Whisper (large) & \href{https://huggingface.co/openai/whisper-large}{openai/whisper-large} & Lokal & Hugging Face & MIT \\
Whisper (v2) & \href{https://huggingface.co/openai/whisper-large-v2}{openai/whisper-large-v3} & Lokal & Hugging Face & MIT \\
Whisper (v3) & \href{https://huggingface.co/openai/whisper-large-v3}{openai/whisper-large-v3} & Lokal & Hugging Face & MIT \\
Faster-Whisper & \href{https://github.com/SYSTRAN/faster-whisper}{large-v3 (CTranslate2)} & Lokal & faster-whisper & MIT \\
WhisperX & \href{https://github.com/m-bain/whisperX}{whisperx-base} & Lokal & whisperx & BSD-2 \\
SeamlessM4T v2 & \href{https://huggingface.co/facebook/seamless-m4t-v2-large}{facebook/seamless-m4t-v2-large} & Lokal & Hugging Face & CC-BY-NC 4.0 \\
Wav2Vec2 XLS-R (DE) & \href{https://huggingface.co/facebook/wav2vec2-large-xlsr-53-german}{facebook/wav2vec2-large-xlsr-53-german} & Lokal & Hugging Face & Apache 2.0 \\
Wav2Vec2 XLS-R (DE) & \href{https://huggingface.co/maxidl/wav2vec2-large-xlsr-german}{maxidl/wav2vec2-large-xlsr-german} & Lokal & Hugging Face & Apache 2.0 \\
Wav2Vec2 XLS-R (DE) & \href{https://huggingface.co/jonatasgrosman/wav2vec2-large-xlsr-53-german}{jonatasgrosman/wav2vec2-xlsr-53-de} & Lokal & Hugging Face & Apache 2.0 \\
Voxtral (Mini) & \href{https://mistral.ai/news/voxtral}{voxtral-small-latest} & API\textsuperscript{*} & Mistral SDK & Apache 2.0 \\
Voxtral (Small) & \href{https://mistral.ai/news/voxtral}{voxtral-mini-latest} & API\textsuperscript{*} & Mistral SDK & Apache 2.0 \\
AssemblyAI & \href{https://www.assemblyai.com}{Universal-1} & API & SDK & Proprietär \\
AssemblyAI (v2) & \href{https://www.assemblyai.com/research/universal-2/}{Universal-2} & API & SDK & Proprietär \\
AssemblyAI (nano) & \href{https://www.assemblyai.com/research/universal-2/}{Universal-nano} & API & SDK & Proprietär \\
Deepgram & \href{https://deepgram.com}{Nova-2} & API & SDK & Proprietär \\ \midrule
\multicolumn{5}{l}{\textit{Englischsprachige Modelle (Referenz)}} \\
\midrule
Whisper (tiny) & \href{https://huggingface.co/openai/whisper-tiny.en}{openai/whisper-tiny.en} & Lokal & Hugging Face & MIT \\
Whisper (base) & \href{https://huggingface.co/openai/whisper-base.en}{openai/whisper-base.en} & Lokal & Hugging Face & MIT \\
Whisper (small) & \href{https://huggingface.co/openai/whisper-small.en}{openai/whisper-small.en} & Lokal & Hugging Face & MIT \\
Whisper (medium) & \href{https://huggingface.co/openai/whisper-medium.en}{openai/whisper-medium.en} & Lokal & Hugging Face & MIT \\
Distil-Whisper (small) & \href{https://huggingface.co/distil-whisper/distil-small.en}{distil-whisper/distil-small.en} & Lokal & Hugging Face & MIT \\
Distil-Whisper (medium) & \href{https://huggingface.co/distil-whisper/distil-medium.en}{distil-whisper/distil-medium.en} & Lokal & Hugging Face & MIT \\
Distil-Whisper (v2) & \href{https://huggingface.co/distil-whisper/distil-large-v2}{distil-large-v2} & Lokal & Hugging Face & MIT \\
Distil-Whisper (v3) & \href{https://huggingface.co/distil-whisper/distil-large-v3}{distil-large-v3} & Lokal & Hugging Face & MIT \\
Wav2Vec2 (base) & \href{https://huggingface.co/facebook/wav2vec2-base-960h}{facebook/wav2vec2-base-960h} & Lokal & Hugging Face & Apache 2.0 \\ 
Wav2Vec2 (large) & \href{https://huggingface.co/facebook/wav2vec2-large-960h}{facebook/wav2vec2-large-960h} & Lokal & Hugging Face & Apache 2.0 \\ 
Wav2Vec2 (large, libri) & \href{https://huggingface.co/facebook/wav2vec2-large-robust-ft-libri-960h}{facebook/wav2vec2-large-robust-ft-libri-960h} & Lokal & Hugging Face & Apache 2.0 \\ 
Wav2Vec2 (large,lv60) & \href{https://huggingface.co/facebook/wav2vec2-large-960h-lv60-self}{facebook/wav2vec2-large-960h-lv60-self} & Lokal & Hugging Face & Apache 2.0 \\ 
\bottomrule
\end{tabular}%
}
\end{table}

\noindent \footnotesize{\textsuperscript{*}Das Modell Voxtral von Mistral AI wurde über die API getestet, ist jedoch unter Apache 2.0 Lizenz auch für den lokalen Betrieb (offene Gewichte) verfügbar.}

\vspace{1em}
\normalsize

Wie in der Auswertung zu sehen sein wird, lieferten die rein englischsprachigen Modelle (Wav2Vec2 EN, Whisper .en) erwartungsgemäß keine brauchbare Transkription der deutschen Anamnesegespräche. Sie dienten primär als Negativkontrolle. Auch das multimodale Modell SeamlessM4T zeigte Schwächen in der spezifischen medizinischen Terminologie im Vergleich zu spezialisierten ASR-Systemen.

\subsection{Evaluations-Metriken}
Zur quantitativen Bestimmung der Transkriptionsgüte und der strukturellen Korrektheit der Sprecherzuordnung werden drei komplementäre Metriken definiert. Die Berechnung basiert primär auf der Wort- bzw. Zeichenebene im Vergleich zu einer manuell verifizierten Ground-Truth-Referenz.

\begin{enumerate}
    \item \textbf{Word Error Rate (WER):} Als primäres Maß für die Erkennungsleistung definiert die WER das Verhältnis der notwendigen Editieroperationen (Substitutionen, Deletionen, Insertionen) zur Gesamtzahl der Wörter in der Referenz. Die Berechnung folgt der Levenshtein-Distanz: $WER = (S + D + I) / N$.
    \item \textbf{Character Error Rate (CER):} Aufgrund der morphologischen Komplexität des Deutschen (insb. Komposita) wird die CER zur feingranularen Analyse herangezogen. Sie ist weniger sensitiv gegenüber geringfügigen Wortbildungsvarianten und erlaubt eine präzisere Bewertung der phonetischen Erkennungsleistung.
    \item \textbf{Speaker-Attributed WER (SA-WER):} Um die Kopplung von ASR und Diarization zu evaluieren, wird die SA-WER eingesetzt. Ein Wort gilt hierbei nur dann als korrekt erkannt, wenn sowohl der lexikalische Inhalt als auch das zugewiesene Sprecher-Label (Speaker-ID) mit der Referenz übereinstimmen.
\end{enumerate}

\section{Ergebnisse}
In den folgenden Abschnitten präsentieren wir die quantitativen und qualitativen Resultate unserer Untersuchung. Wir analysieren zunächst die reine Transkriptionsleistung anhand der Wortfehlerrate (WER) sowie der Zeichenfehlerrate (CER), betrachten anschließend die Qualität der Sprechertrennung und diskutieren abschließend die Stabilität der Modelle über verschiedene akustische Szenarien hinweg.

\subsection{Ergebnisse der Transkriptionsanalyse}
Die Evaluation zeigt eine deutliche Korrelation zwischen der Modellarchitektur (Parameteranzahl, Trainingsdaten-Diversität) und der resultierenden Wortfehlerrate. Tabelle \ref{tab:wer_results} fasst die aggregierten Ergebnisse der führenden Modelle zusammen.

\begin{itemize}
    \item \textbf{Hochperformante API-Lösungen:} Das Modell \textit{AssemblyAI Universal} emittiert mit einer WER von $3,0\,\%$ die geringste Fehlerrate im Testfeld. Die geringe Varianz über alle Datensätze hinweg deutet auf eine hohe Robustheit der zugrunde liegenden \textit{Conformer}-Architektur hin.
    \item \textbf{Effizienz-Analyse (Mistral Voxtral):} Die Modelle der \textit{Voxtral}-Serie zeigen eine kompetitive Performance. Insbesondere \textit{Voxtral Small} erreicht eine WER von $7,1\,\%$, was im Kontext der Modellgröße eine hohe Effizienz pro Parameter signalisiert.
    \item \textbf{Open-Source Benchmarks:} \textit{OpenAI Whisper Large-v3} erreicht eine WER von $12,6\,\%$. Qualitativ ist hierbei eine Tendenz zu Fehlern bei akustisch schwach repräsentierten Segmenten festzustellen, was die Notwendigkeit von Post-Processing-Schritten unterstreicht.
    \item \textbf{Degradierung bei Legacy-Modellen:} Ältere Architekturen (z.\,B. \textit{Wav2Vec2}) sowie kompakte Whisper-Varianten zeigen Fehlerraten von $> 20\,\%$, was für eine automatisierte medizinische Dokumentation vermutlich als unzureichend eingestuft werden kann.
\end{itemize}

\clearpage
\begin{table}[ht!]
    \centering
    \small 
    \caption{Selektiver Vergleich der ASR-Performance}
    \label{tab:wer_results}
    \vspace{5pt}
    \begin{tabular}{lccc} 
        \toprule
        \textbf{Modell} & \textbf{WER ($\varnothing$)} & \textbf{CER ($\varnothing$)} & \textbf{BLEU} \\
        \midrule
        AssemblyAI Universal & 0.0299 & 0.0194 & 0.942 \\
        Voxtral Small Latest & 0.0711 & 0.0355 & 0.866 \\
        Voxtral Mini Latest  & 0.0846 & 0.0389 & 0.842 \\
        Whisper Large-v3 (Distilled) & 0.1258 & 0.0809 & 0.817 \\
        Deepgram Nova-2      & 0.1433 & 0.0742 & 0.756 \\
        WhisperX-Base        & 0.2151 & 0.1033 & 0.663 \\
        \bottomrule
    \end{tabular}
\end{table}

\subsection{Ergebnisse im Detail und Modellvergleich}

\begin{figure}[htbp]
    \centering
    \includegraphics[width=\linewidth]{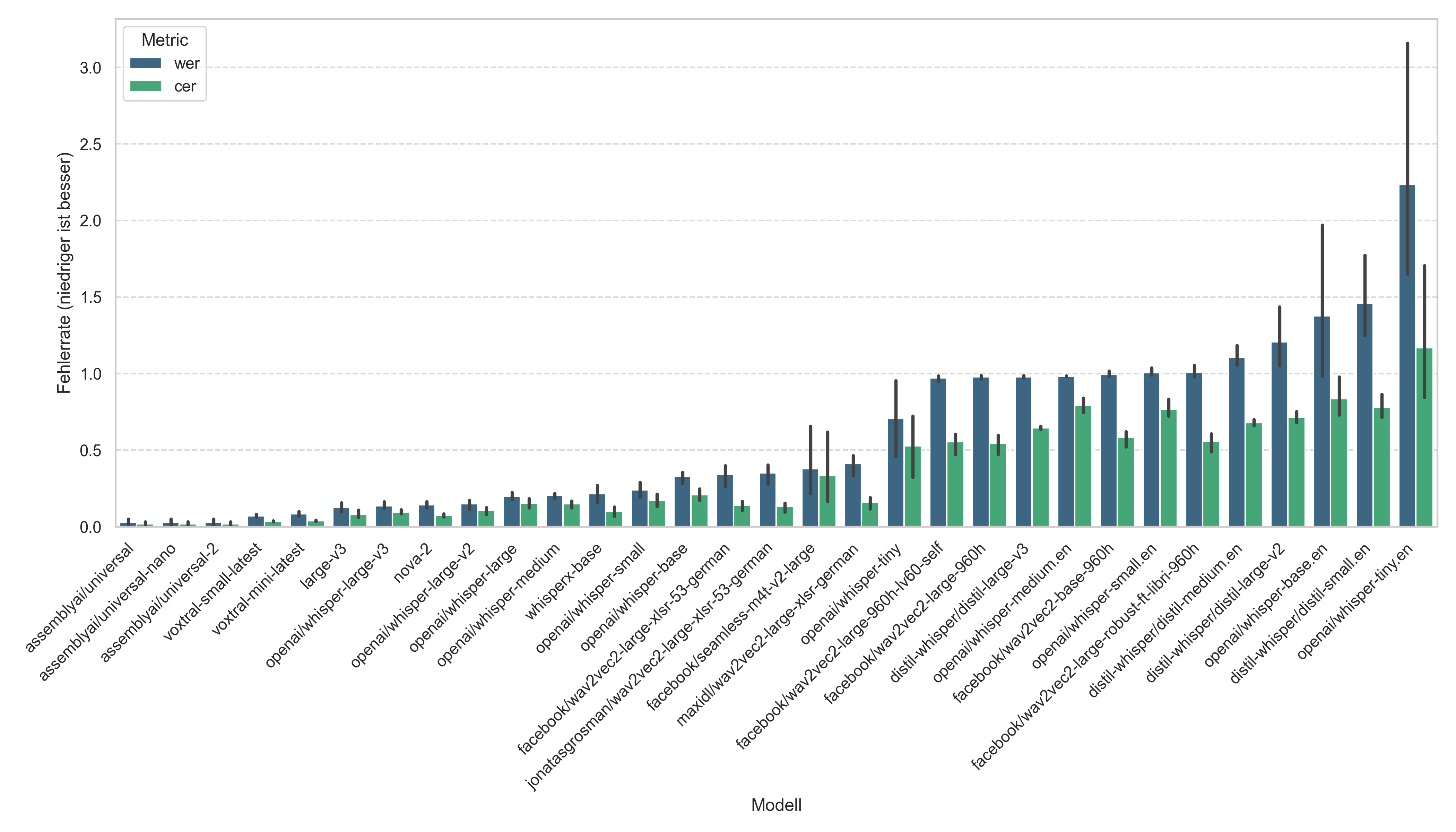}
    \caption{Durchschnittliche WER und CER je Modell (Fulltext-Auswertung).}
    \label{fig:bar_wer_cer}
\end{figure}

Die Gegenüberstellung der durchschnittlichen Wortfehlerraten (WER) und Zeichenfehlerraten (CER) über das gesamte Test-Korpus (siehe Abb. \ref{fig:bar_wer_cer}) lässt eine klare Stratifizierung des Testfeldes erkennen. Während aktuelle SOTA-Modelle (State-of-the-Art) bereits Fehlerraten im teils niedrigen einstelligen Prozentbereich erreichen, weisen besonders kleinere sowie ältere Open-Source-Derivate eine signifikante Distanz zur geforderten Transkriptionsgüte auf.

\begin{figure}[htbp]
    \centering
    \includegraphics[width=\linewidth]{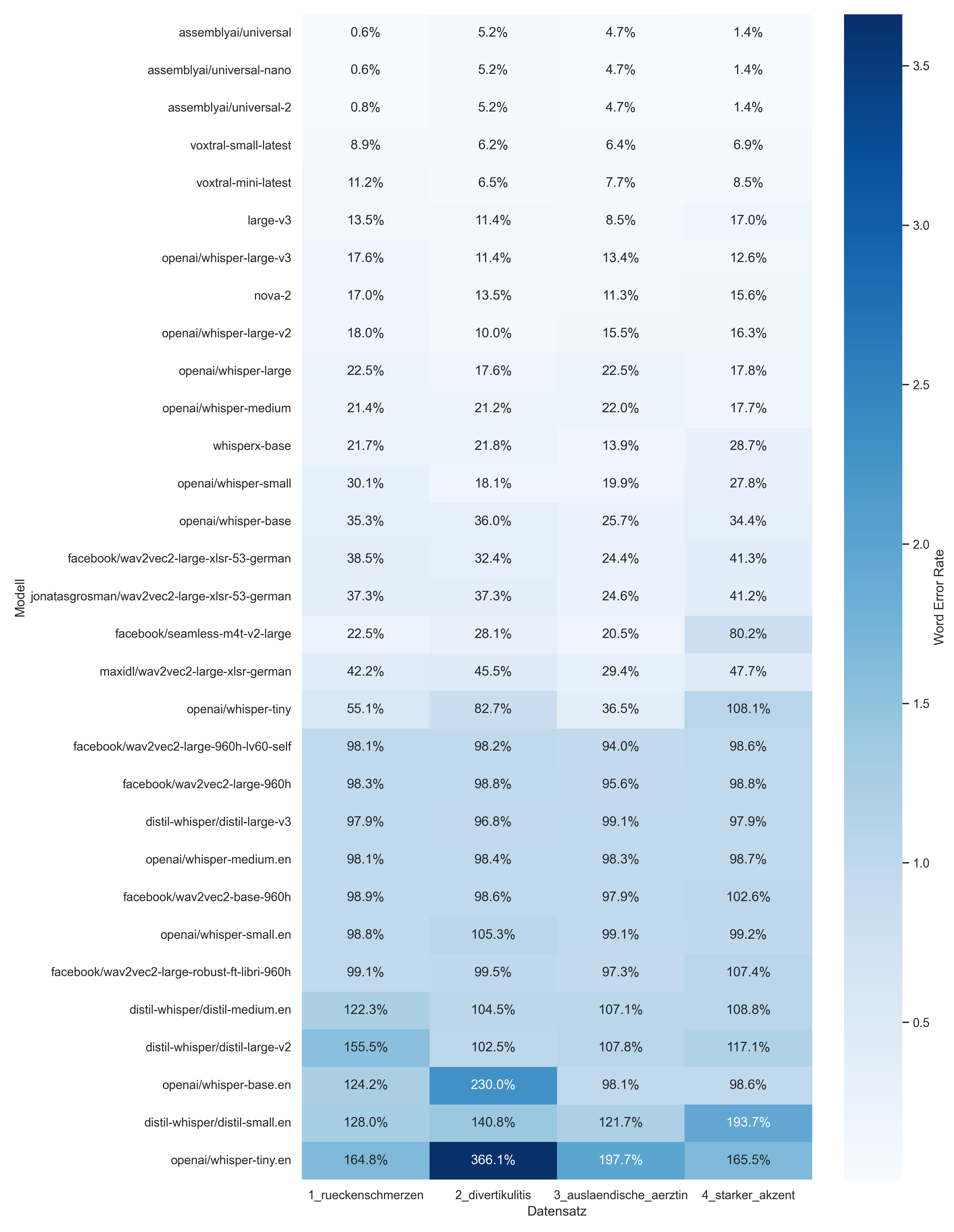}
    \caption{Word-Error-Rate Heatmap über alle Modelle und pro Datensatz}
    \label{fig:heatmap}
\end{figure}

\noindent
\textbf{Kontextabhängige Robustheit:} 
Die detaillierte Heatmap-Analyse in Abbildung \ref{fig:heatmap} verdeutlicht die Fehlervarianz in Abhängigkeit des gewählten Szenarios. Es zeigt sich, dass insbesondere akustisch herausfordernde Umgebungen sowie domänenspezifische Fachterminologie als primäre Fehlertreiber fungieren. Hochperformante Modelle wie \textit{AssemblyAI Universal} und \textit{Mistral's Voxtral}demonstrieren hier eine bemerkenswerte Invarianz gegenüber diesen Störfaktoren, während bei Modellen wie \textit{Whisper Large-v3} die Fehlerrate bei komplexen Akzenten deutlich ansteigt.

\begin{figure}[htbp]
    \centering
    \includegraphics[width=0.9\linewidth]{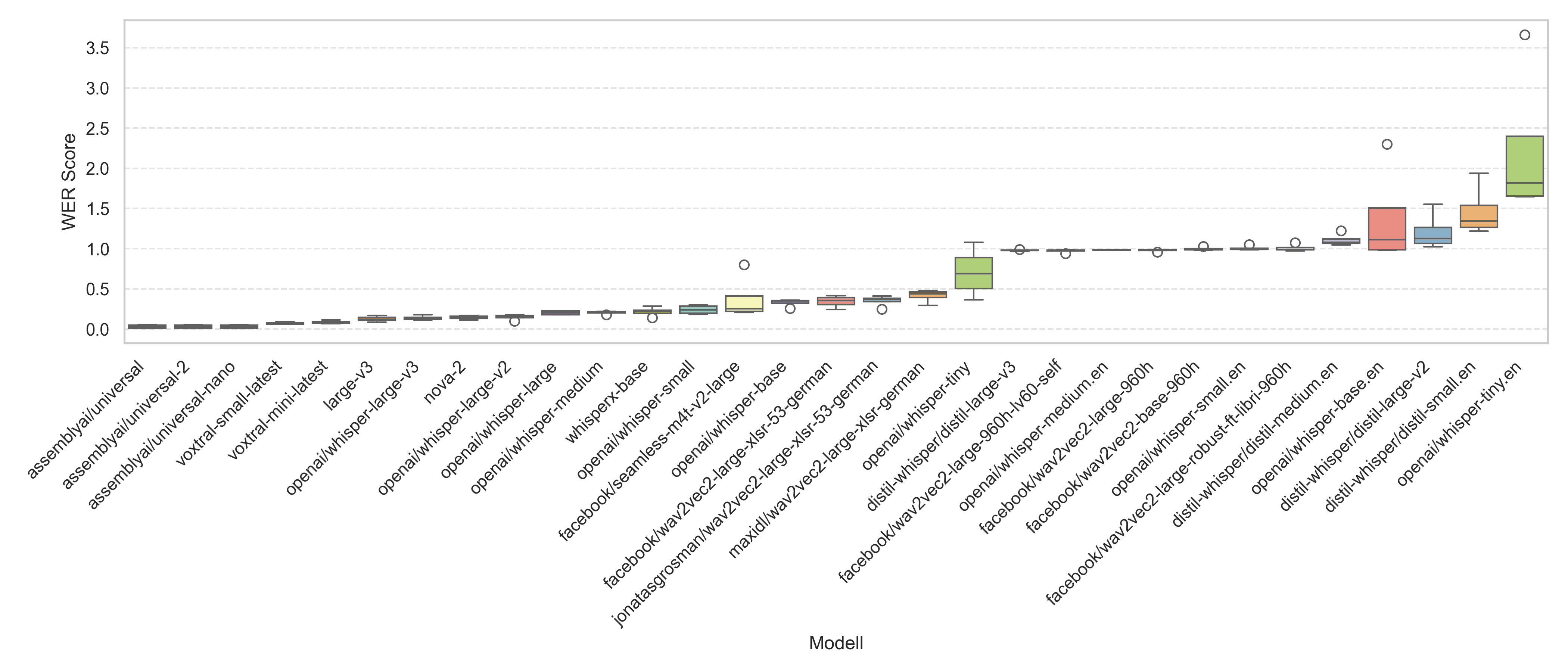}
    \caption{Verteilung der WER pro Modell (Stabilitätsanalyse).}
    \label{fig:box_stability}
\end{figure}

\begin{figure}[htbp]
    \centering
    \includegraphics[width=0.9\linewidth]{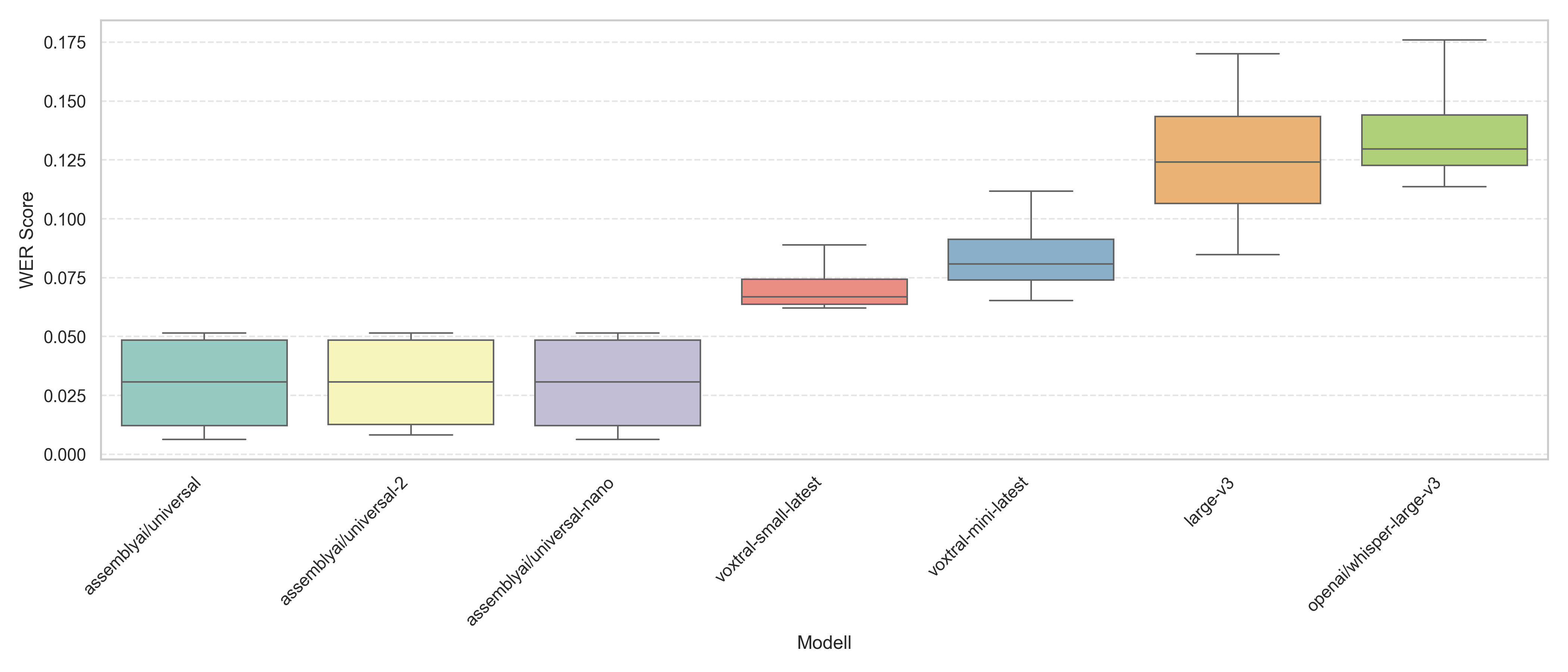}
    \caption{Verteilung der WER für die 7 besten Modelle (Stabilitätsanalyse). }
    \label{fig:box_stability_top7}
\end{figure}

\noindent
\textbf{Statistische Stabilität und Zuverlässigkeit:} 
Für den produktiven Einsatz im medizinischen Umfeld ist die Vorhersagbarkeit der Performance essenziell. Die Stabilitätsanalyse in Abbildung \ref{fig:box_stability} offenbart bei Modellen mit hoher durchschnittlicher WER eine gleichzeitig hohe Varianz und zahlreiche Ausreißer. Eine Detailbetrachtung der sieben leistungsstärksten Modelle (Abb. \ref{fig:box_stability_top7}) belegt jedoch, dass die Spitzenreiter nicht nur präzise, sondern auch statistisch stabil agieren. Dies minimiert das Risiko von unvorhersehbaren Halluzinationen in der medizinischen Dokumentation.

\begin{figure}[htbp]
    \centering
    \includegraphics[width=0.7\linewidth]{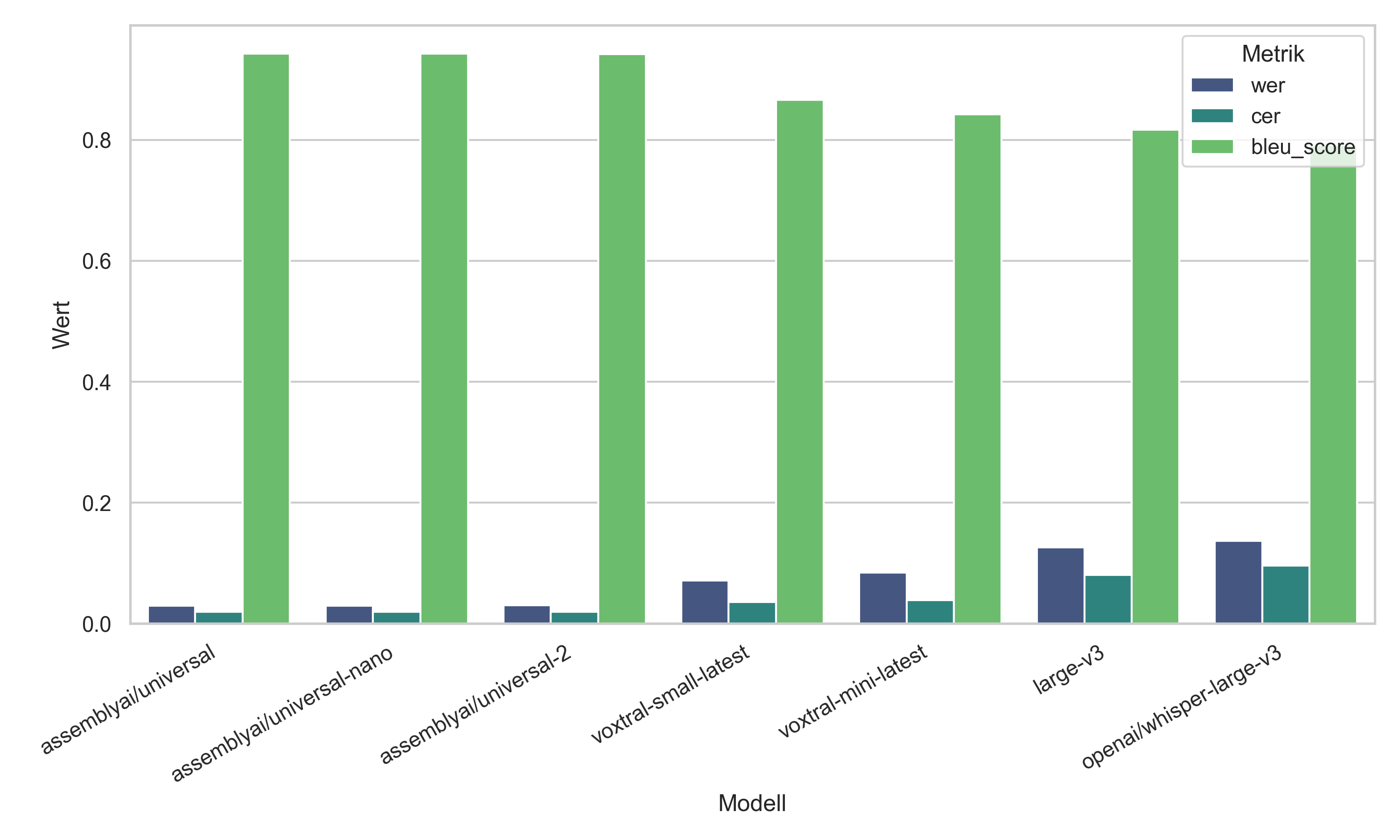}
    \caption{Vergleich der Top-7-Modelle nach WER: WER, CER und BLEU-Score.}
    \label{fig:top7}
\end{figure}

\clearpage

\noindent
\textbf{Semantische und linguistische Integrität:} 
Der Vergleich der Top-7-Modelle in Abbildung \ref{fig:top7} integriert neben der WER und CER auch den \textit{BLEU-Score} (Bilingual Evaluation Understudy). Es ist eine starke positive Korrelation zwischen niedrigen Fehlerraten und hohen BLEU-Werten zu beobachten (Bestwert $> 0,94$). Dies indiziert, dass die Transkripte der führenden Modelle nicht nur lexikalisch korrekt sind, sondern die ursprüngliche semantische Struktur und den Satzbau der Probanden nahezu verlustfrei abbilden. Diese linguistische Präzision ist die Grundvoraussetzung für die nachgelagerte automatisierte Analyse durch LLM-basierte Assistenzsysteme.

\subsection{Sprechererkennung (Diarization)}
Die Fähigkeit zur Unterscheidung von Arzt und Patient wurde separat evaluiert.
\begin{itemize}
    \item \textbf{Deepgram Nova-2:} Zeigte eine solide Leistung in der Sprechertrennung \cite{deepgram2024nova}, litt jedoch unter einer etwas höheren Basis-WER.
    \item \textbf{AssemblyAI:} Konnte durch die sehr hohe Transkriptionsgenauigkeit auch in der attributierten Auswertung punkten, da weniger Wörter falsch zugeordnet wurden.
    \item \textbf{Open Source:} Viele Open-Source-Wrapper (z.B. Standard HuggingFace-Pipelines) bieten keine integrierte Diarization, was den Einsatz komplexer macht. Hier sind Zusatzmodule wie z.\,B. Pyannote.audio \cite{PyannotePyannoteaudio2026, Plaquet23} notwendig.
\end{itemize}

\begin{figure}[htbp]
    \centering
    \includegraphics[width=0.7\linewidth]{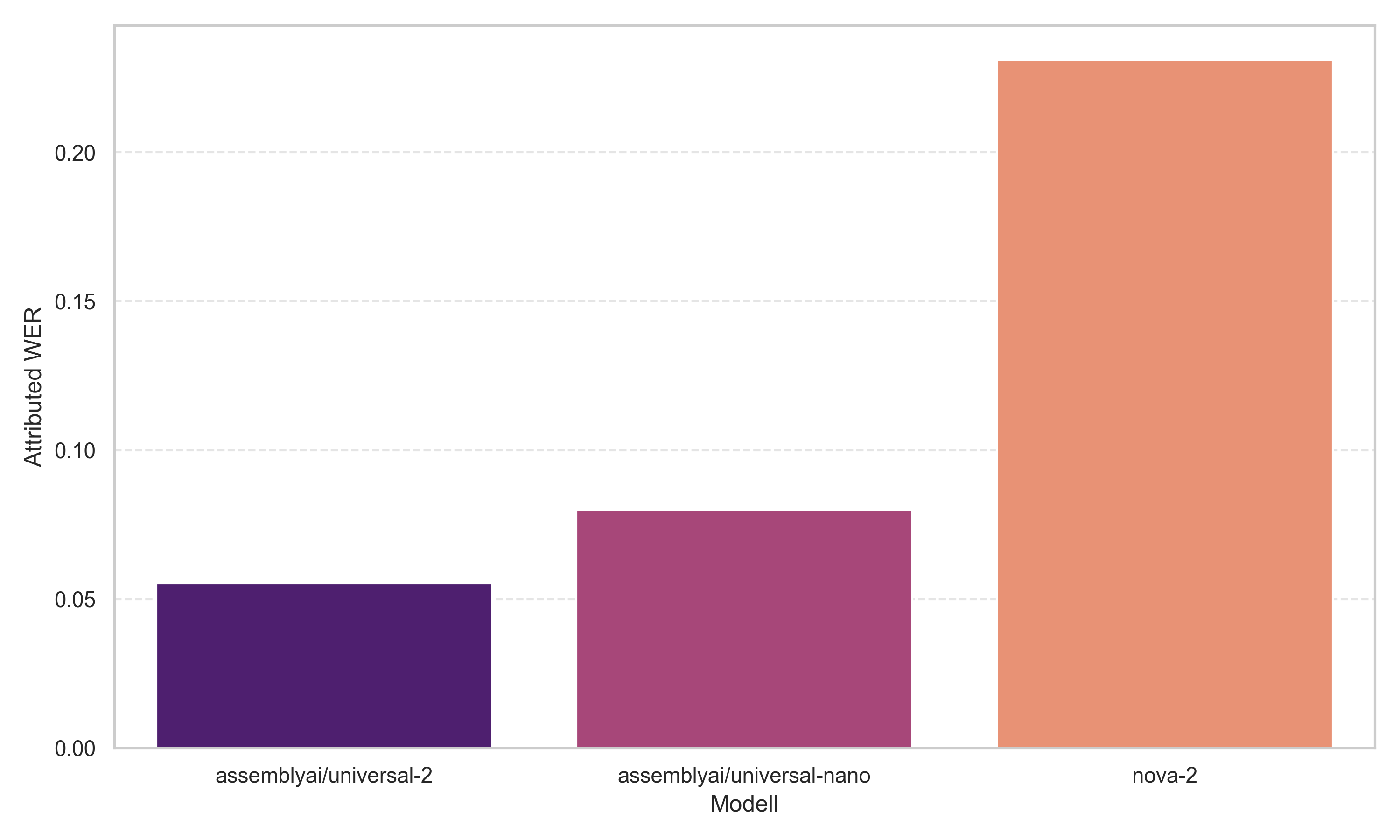}
    \caption{Attributed WER (Sprecher-Zuordnung) für Modelle mit Diarization-Unterstützung}
    \label{fig:diarization}
\end{figure}

\noindent
Aus Abbildung~\ref{fig:diarization} wird ersichtlich, dass ohne Erweiterung nur die kommerziellen Modelle attributierte Transkripte direkt liefern. In unseren Messungen zeigte \textit{universal-2} von AssemblyAI die konsistentesten Ergebnisse in Bezug auf Sprecherzuordnung.

\section{Diskussion}
Die vorliegenden Ergebnisse belegen, dass die neueste Generation spezialisierter ASR-Modelle eine kritische Schwelle für den Einsatz in hochsensiblen medizinischen Domänen erreicht hat.

\begin{itemize}
    \item \textbf{Modell-Souveränität und Datenschutz:} Während kommerzielle APIs wie \textit{AssemblyAI} die quantitative Benchmark anführen, demonstriert insbesondere \textit{Mistral Voxtral} (OpenWeights), dass auch selbstverwaltete Lösungen (bspw. On-Premise) eine nahezu äquivalente Performanz bieten. Für den Einsatz im Rahmen der DSGVO-konformen Verarbeitung stellt dies einen entscheidenden Vorteil dar, da die Datenhoheit gewahrt bleibt, ohne signifikante Einbußen in der Transkriptionsgüte hinnehmen zu müssen.
    \item \textbf{Semantische Informationstreue:} Erste Pilotuntersuchungen mittels einer \textit{LLM-as-a-Judge}-Bewertung, basierend auf einem Vergleich von Ground-Truth-Zusammenfassungen und Transkript-Extraktionen, zeigen für die Top-7-Modelle bereits sehr gute Bewertungen. Die Analyse deutet darauf hin, dass der automatisierte \enquote{Judge} aufgrund struktureller Divergenzen in den LLM-Zusammenfassungen im Hinblick auf die Transkription tendenziell zu streng urteilt, da die Struktur durch das LLM stärker korrigiert wird als der Inhalt. Für die medizinische Dokumentation bleiben die essenziellen Informationen jedoch bereits im Null-Shot-Szenario, also ohne aufwendiges Fine-Tuning, nahezu vollständig erhalten bleiben. Dies reduziert die Notwendigkeit für ressourcenintensive Anpassungsprozesse.
    \item \textbf{Robustheit gegenüber Fachjargon:} Die Korrelation zwischen niedrigen WER-Werten und hohen semantischen Bewertungen bei \textit{AssemblyAI} und \textit{Voxtral} lässt darauf schließen, dass die Modelle über ein ausgeprägtes implizites Wissen über medizinische Kontexte verfügt. 
\end{itemize}

\section{Fazit und Ausblick}
Die Evaluation zeigt, dass automatisierte Bewertungen und Protokollierung bereits mit aktuellen ASR-Modellen möglich sind. Hierbei definieren \textit{AssemblyAI Universal} (Cloud API) und \textit{Mistral Voxtral} (OpenWeights) aktuell die technologische Leistungsspitze. Aufgrund der Anforderungen an die Datensicherheit im medizinischen Umfeld und der bereits exzellenten Out-of-the-box-Performanz verspricht besonders \textit{Voxtral} eine sehr gute Basis zu sein, um konkrete Systeme umzusetzen.

In zukünftigen Forschungsarbeiten werden wir die semantische Analyseebene weiter vertiefen. Hierbei stehen drei Schwerpunkte im Fokus:

\begin{enumerate}
    \item \textbf{Hybrid-Evaluation:} Die Kombination aus automatisierten Bwerbungsverfahren (\textit{LLM-as-a-Judge}) und systematischer menschlicher Bewertung (\textit{Human Evaluation}) zur Validierung der klinischen Relevanz. Ziel ist es, die Diskrepanz zwischen rein statistischen Fehlerraten (WER) und dem tatsächlichen Informationsverlust zu quantifizieren.
    \item \textbf{Domänenspezifische Fehlermetriken:} Nutzung einer \textit{Keyword-specific Word Error Rate} (kWER), die die Erkennungsrate explizit auf das medizinische Fachvokabular bezieht. Parallel dazu soll eine \textit{Human-centered Word Error Rate} (hWER) die Lesbarkeit und den Nachbearbeitungsaufwand für das medizinische Personal evaluieren.
    \item \textbf{Semantische Extraktion:} Die Weiterentwicklung von Agenten-Systemen, die direkt aus den stabilen Transkripten strukturierte Datenformate (z.\,B. nach FHIR-Standard) extrahieren, um die Interoperabilität mit bestehenden Kliniksystemen zu gewährleisten.
\end{enumerate}

\bibliographystyle{plain}
\bibliography{references}

\end{document}